%% file: main.tex
\documentclass[dvipsnames]{article} 
\usepackage{iclr2021_conference,times}

\input{math_commands.tex}

\usepackage{hyperref}
\usepackage{url}
\usepackage{titlecaps}
\usepackage{booktabs}
\usepackage[ruled,vlined]{algorithm2e}
\usepackage{nicefrac}
\usepackage{graphicx}
\usepackage{subcaption}
\usepackage{booktabs}
\usepackage{amsthm}
\usepackage{bbm}

\title{Viewmaker Networks: Learning Views for \\ Unsupervised Representation Learning}

\author{Alex Tamkin, Mike Wu, Noah Goodman \\
Department of Computer Science\\
Stanford University \\
Stanford, CA 94305, USA \\
\texttt{\{atamkin, wumike, ngoodman\}@stanford.edu} \\
}

\usepackage{amsmath}
\usepackage{xcolor}

\PassOptionsToPackage{hyphens}{url}\usepackage{hyperref}

\iclrfinalcopy 
\begin{document}

\newcommand{\ours}{viewmaker}

\maketitle

\begin{abstract}
\input{abstract.tex}
\end{abstract}

\input{intro.tex}
\input{related.tex}
\input{method.tex}

\input{images.tex}
\input{audio.tex}
\input{imu.tex}
\input{conclusion.tex}
\input{ack.tex}

\bibliography{iclr2021_conference}
\bibliographystyle{iclr2021_conference}

\appendix
\input{appendix.tex}

\end{document}

%% file: math_commands.tex
\usepackage{amsmath,amsfonts,bm}

\def\eqref#1{equation~\ref{#1}}

\def\1{\bm{1}}

\def\eps{{\epsilon}}

\DeclareMathAlphabet{\mathsfit}{\encodingdefault}{\sfdefault}{m}{sl}
\SetMathAlphabet{\mathsfit}{bold}{\encodingdefault}{\sfdefault}{bx}{n}



%% file: abstract.tex
Many recent methods for unsupervised representation learning train models to be invariant to different ``views,'' or distorted versions of an input. However, designing these views requires considerable trial and error by human experts, hindering widespread adoption of unsupervised representation learning methods across domains and modalities. 
To address this, we propose \emph{viewmaker networks}: generative models that learn to produce useful views from a given input.
Viewmakers are \emph{stochastic bounded adversaries}: they produce views by generating and then adding an $\ell_p$-bounded perturbation to the input, and are trained adversarially with respect to the main encoder network.
Remarkably, when pretraining on CIFAR-10, our learned views enable comparable transfer accuracy to the well-tuned SimCLR augmentations---despite not including transformations like cropping or color jitter. Furthermore, our learned views significantly outperform baseline augmentations on speech recordings (+9\% points, on average) and wearable sensor data (+17\% points).
Viewmakers can also be combined with handcrafted views: they improve robustness to common image corruptions and can increase transfer performance in cases where handcrafted views are less explored. 
These results suggest that viewmakers may provide a path towards more general representation learning algorithms---reducing the domain expertise and effort needed to pretrain on a much wider set of domains. Code is available at \url{https://github.com/alextamkin/viewmaker}.

%% file: intro.tex
\begin{figure}[h]
    \centering
    \includegraphics[width=95pt]{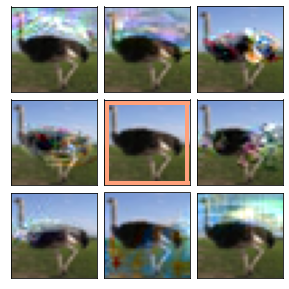}
    \includegraphics[width=95pt]{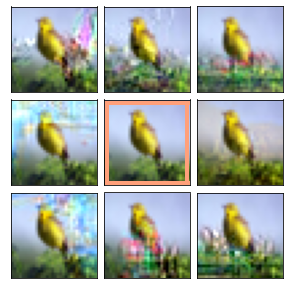}
    \includegraphics[width=95pt]{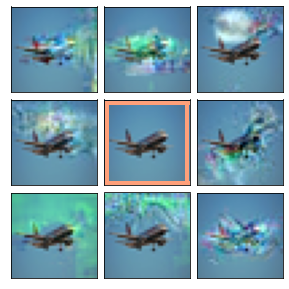}
    \includegraphics[width=95pt]{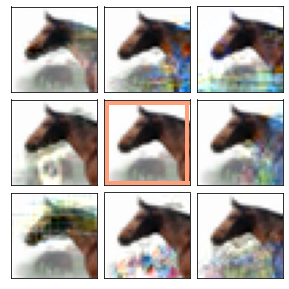}
    \caption{\textbf{\titlecap{\ours{}} networks generate complex and diverse input-dependent views for unsupervised learning.} Examples shown are for CIFAR-10. Original image in center with pink border.}
    \label{fig:cifar10-views}
\end{figure}

\section{Introduction}

Unsupervised representation learning has made significant recent strides, including in computer vision, where view-based methods have enabled strong performance on benchmark tasks \citep{wu2018unsupervised, oord2018representation, bachman2019learning, zhuang2019local, misra2020self, he2020momentum, chen2020simple}. \emph{Views} here refer to human-defined data transformations, which target capabilities or invariances thought to be useful for transfer tasks. In particular, in contrastive learning of visual representations, models are trained to maximize the mutual information between different views of an image, including crops, blurs, noise, and changes to color and contrast \citep{bachman2019learning, chen2020simple}. Much work has investigated the space of possible image views (and their compositions) and understanding their effects on transfer learning \citep{chen2020simple, wu2020mutual, tian2019contrastive, purushwalkam2020demystifying}.

The fact that views must be designed by hand is a significant limitation. While views for image classification have been refined over many years, new views must be developed from scratch for new modalities. Making matters worse, even \emph{within} a modality, different domains may have different optimal views \citep{purushwalkam2020demystifying}. Previous studies have investigated the properties of good views through the lens of mutual information \citep{tian2020makes, wu2020mutual}, but a broadly-applicable approach for learning views remains unstudied.

In this work, we present a general method for learning diverse and useful views for contrastive learning. Rather than searching through possible compositions of existing view functions \citep{cubuk2018autoaugment, lim2019fast}, which may not be available for many modalities, our approach produces views with a generative model, called a \emph{\ours{}} network, trained jointly with the encoder network. This flexibility enables learning a broad set of possible view functions, including input-dependent views, without resorting to hand-crafting or expert domain knowledge.
The \ours{} network is trained adversarially to create views which increase the contrastive loss of the encoder network. Rather than directly outputting views for an image, the \ours{} instead outputs a stochastic perturbation that is \emph{added} to the input. This perturbation is projected onto an $\ell_p$ sphere, controlling the effective strength of the view, similar to methods in adversarial robustness. This constrained adversarial training method enables the model to reduce the mutual information between different views while preserving useful input features for the encoder to learn from.

In summary, we contribute:
\begin{enumerate}
    \item \titlecap{\ours{}} networks: to our knowledge the first modality-agnostic method to \emph{learn} views for unsupervised representation learning
    \item On image data, where expert-designed views have been extensively optimized, our \ours{}-models achieve comparable transfer performance to state of the art contrastive methods while being more robust to common corruptions.
    \item On speech data, our method significantly outperforms existing human-defined views on a range of speech recognition transfer tasks.
    \item On time-series data from wearable sensors, our model significantly outperforms baseline views on the task of human activity recognition (e.g., cycling, running, jumping rope). 
\end{enumerate}

%% file: related.tex
\section{Related work}

\paragraph{Unsupervised representation learning} Learning useful representations from unlabeled data is a fundamental problem in machine learning \citep{pan2009survey, bengio2013representation}. A recently successful framework for unsupervised representation learning for images involves training a model to be invariant to various data transformations \citep{bachman2019learning, misra2020self}, although the idea has much earlier roots \citep{becker1992self, hadsell2006dimensionality, dosovitskiy2014discriminative}. This idea has been expanded by a number of contrastive learning approaches which push embeddings of different views, or transformed inputs, closer together, while pushing other pairs apart \citep{tian2019contrastive,he2020momentum,chen2020simple,chen2020big,chen2020improved}, as well as non-contrastive approaches which do not explicitly push apart unmatched views \citep{grill2020bootstrap, caron2020unsupervised}. Related but more limited setups have been explored for speech, where data augmentation strategies are less explored \citep{oord2018representation, kharitonov2020data}.

\paragraph{Understanding and designing views} Several works have studied the role of views in contrastive learning, including from a mutual-information perspective \citep{wu2020mutual}, in relation to specific transfer tasks \citep{tian2019contrastive}, with respect to different kinds of invariances \citep{purushwalkam2020demystifying}, or via careful empirical studies \citep{chen2020simple}. Outside of a contrastive learning framework, \citet{gontijo2020affinity} study how data augmentation aids generalization in vision models. Much work has explored different handcrafted data augmentation methods for supervised learning of images \citep{hendrycks2020many, lopes2019improving, perez2017effectiveness, yun2019cutmix, zhang2017mixup}, speech \citep{park2019specaugment, kovacs2017increasing, toth2018perceptually, kharitonov2020data}, or in feature space \citep{devries2017dataset}.

\paragraph{Adversarial methods} Our work is related to and inspired by work on adversarial methods, including the $\ell_p$ balls studied in adversarial robustness \citep{szegedy2013intriguing, madry2017towards, raghunathan2018certified} and training networks with adversarial objectives \citep{goodfellow2014generative, xiao2018generating}. Our work is also connected to the vicinal risk minimization principle \citep{chapelle2001vicinal} and can be interpreted as producing amortized virtual adversarial examples \citep{miyato2018virtual}. Previous adversarial view-based pretraining methods add adversarial noise on top of existing handcrafted views \citep{kim2020adversarial} or require access to specific transfer tasks during pretraining \citep{tian2020makes}. In contrast, our method is more general: it is neither specialized to a particular downstream task, nor requires neither human-defined view families. Outside of multi-view learning paradigms, adversarial methods have also seen use for representation learning in GANs \citep{donahue2016adversarial, NIPS2019_9240} or in choosing harder negative samples \citep{bose2018adversarial}, as well as for data augmentation \citep{antoniou2017data, volpi2018generalizing, bowles2018gan}. Adversarial networks that perturb inputs have also been investigated to improve GAN training \citep{sajjadi2018tempered} and to remove ``shortcut'' features (e.g., watermarks) for self-supervised pretext tasks \citep{minderer2020automatic}. 

\paragraph{Learning views} Outside of adversarial approaches, our work is related to other studies that seek to learn data augmentation strategies by composing existing human-designed augmentations \citep{ratner2017learning, cubuk2018autoaugment, zhang2019adversarial, ho2019population, lim2019fast, cubuk2020randaugment} or by modeling variations specific to the data distribution \citep{tran2017bayesian, wong2020learning}. By contrast, our method requires no human-defined view functions, does not require first pretraining a generative model, and can generate perturbations beyond naturally-occurring variation observed in the training data (e.g. brightness or contrast), potentially conferring robustness benefits, as we explore in Section \ref{subsec:cifar-c}.

%% file: method.tex
\section{Method}

\begin{figure}
    \centering
    \includegraphics[width=\textwidth]{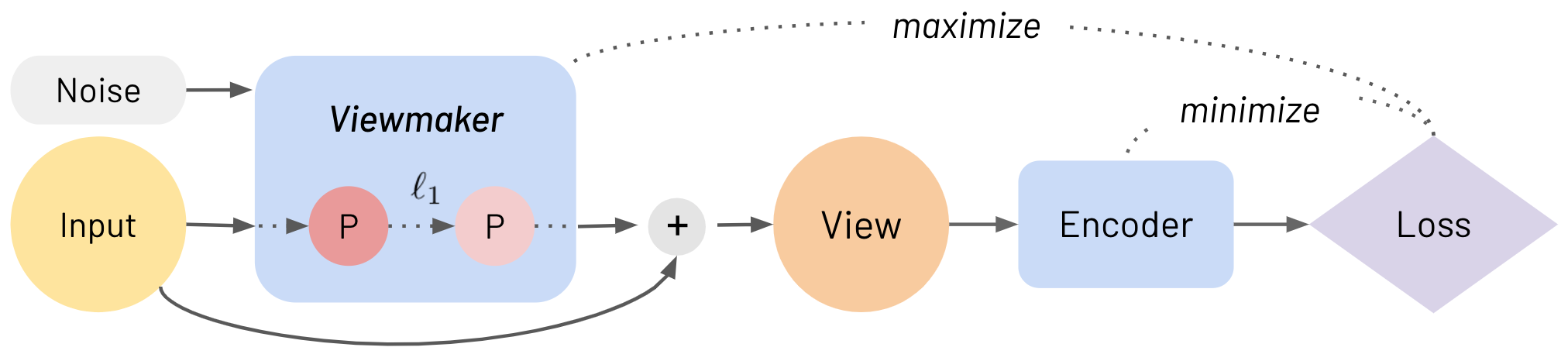}
    \caption{\textbf{Diagram of our method.} The \ours{} network is trained to produce stochastic adversarial views restricted to an $\ell_1$ sphere around the input.}
    \label{fig:vm-fig}
\end{figure}

In contrastive learning, the objective is to push embeddings of positive views (derived from the same input) close together, while pushing away embeddings of negative views (derived from different inputs). We focus mainly on the simple, yet performant, SimCLR contrastive learning algorithm \citep{chen2020simple}, but we also consider a memory bank-based algorithm \citep{wu2018unsupervised} in Section \ref{sec:images}. As our method is agnostic to the specific pretraining loss used, it is naturally compatible with other view-based algorithms such as MoCo \citep{he2020momentum}, BYOL \citep{grill2020bootstrap}, and SwAV \citep{caron2020unsupervised} by similarly substituting the data transformation pipeline with a \ours{} network.

Formally, given a batch of $N$ pairs of positive views $(i,j)$ the SimCLR loss is 
\begin{align*}
    \mathcal{L} = \frac{1}{2N} \sum_{k=1}^N [\ell(2k-1, 2k) + \ell(2k, 2k-1)] 
    \text{\; where \;}
    \ell(i, j) = -\log \frac{\exp(s_{i,j} / \tau)}{\sum_{k=1}^{2N} \mathbbm{1}_{[k \neq i]} \exp(s_{i,k}/\tau)}
\end{align*}
and $s_{a,b}$ is the cosine similarity of the embeddings of views $a$ and $b$.

We generate views by perturbing examples with a \emph{\ours} network $V$, trained jointly with the main \emph{encoder} network $M$. There are three attributes desirable for useful perturbations, each of which motivates an aspect of our method:

\begin{enumerate}
    \item \textbf{Challenging:} The perturbations should be complex and strong enough that an encoder must develop useful representations to perform the self-supervised task. We accomplish this by generating perturbations with a neural network that is trained adversarially to increase the loss of the encoder network. Specifically, we use a neural network that ingests the input $X$ and outputs a view $X + V(X)$.
    \item \textbf{Faithful:} The perturbations must not make the encoder task impossible, being so strong that they destroy all features of the input. For example, perturbations should not be able to zero out the input, making learning impossible. We accomplish this by constraining the perturbations to an $\ell_p$ sphere around the original input.
    $\ell_p$ constraints are common in the adversarial robustness literature where perturbations are expected to be indistinguishable. 
    In our experiments, we find the best results are achieved with an $\ell_1$ sphere, which grants the \ours{} a \emph{distortion budget} that it can spend on a small perturbation for a large part of the input or a more extreme perturbation for a smaller portion.
    \item \textbf{Stochastic:} The method should be able to generate a variety of perturbations for a single input, as the encoder objective requires contrasting two different views of an input against each other. To do this, we inject random noise into the \ours{}, such that the model can learn a stochastic function that produces a different perturbed input each forward pass.
\end{enumerate}

Figure \ref{fig:vm-fig} summarizes our method. The encoder and \ours{} are optimized in alternating steps to minimize and maximize $\mathcal L$, respectively. We use an image-to-image neural network as our \ours{} network, with an architecture adapted from work on style transfer \citep{10.1007/978-3-319-46475-6_43}. See the Appendix for more details. This network ingests the input image and outputs a perturbation that is constrained to an $\ell_1$ sphere. The sphere's radius is determined by the volume of the input tensor times a hyperparameter $\epsilon$, the \emph{distortion budget}, which determines the strength of the applied perturbation. This perturbation is added to the input image and optionally clamped in the case of images to ensure all pixels are in $[0,1]$. Algorithm \ref{alg:view-generation} describes this process precisely.

\newcommand\mycommfont[1]{\footnotesize\ttfamily\textcolor{blue}{#1}}
\SetCommentSty{mycommfont}
\begin{algorithm}[H]
\SetAlgoLined
\KwIn{\titlecap{\ours{}} network $V$, $C\times W\times H$ image X, $\ell_1$ distortion budget $\epsilon$, noise $\delta$}
\KwOut{Perturbed $C\times W\times H$ image $X$}
 $P \gets V(X, \delta)$ \tcp{generate perturbation}
 $P \gets \frac{\epsilon CWH}{|P|_1} P$\ \tcp{project to $\ell_1$ sphere}
 $X \gets X + P$\ \tcp{apply perturbation}
 $X \gets \text{clamp}(X, 0, 1)$ \tcp{clamp (images only)}
 \caption{Generating \ours{} views}
 \label{alg:view-generation}
\end{algorithm}

%% file: images.tex
\section{Images}
\label{sec:images}

We begin by applying the \ours{} to contrastive learning for images. In addition to SimCLR \citep{chen2020simple}, we also consider a memory bank-based instance discrimination framework \citep[][henceforth InstDisc]{wu2018unsupervised}.

We pretrain ResNet-18 \citep{he2015deep} models on CIFAR-10 \citep{Krizhevsky09learningmultiple} for 200 epochs with a batch size of 256. We train a \ours{}-encoder system  with a distortion budget of $\epsilon=0.05$.  We tried distortion budgets $\epsilon \in \{0.1, 0.05, 0.02\}$ and found $0.05$ to work best; however, we anticipate that further tuning would yield additional gains. As we can see in Figure \ref{fig:cifar10-views}, the learned views are diverse, consisting of qualitatively different kinds of perturbations and affecting different parts of the input. We compare the resulting encoder representations with a model trained with the \emph{expert views} used for SimCLR, comprised of many human-defined transformations targeting different kinds of invariances useful for image classification: cropping-and-resizing, blurring, horizontal flipping, color dropping, and shifts in brightness, contrast, saturation, and hue \citep{chen2020simple}.

\subsection{Transfer results on image classification tasks}

\begin{table}[h!]
    \centering
    \begin{tabular}{lrrrr}
        \toprule
         & \multicolumn{2}{c}{SimCLR} & \multicolumn{2}{c}{InstDisc}  \\
        \cmidrule(lr){2-3}
        \cmidrule(lr){4-5}
         Dataset & Expt & Ours & Expt & Ours \\
         \midrule
         CIFAR-10 & $\mathbf{86.2}$ & $84.5$ & $\mathbf{82.4}$ & $80.1$ \\
         MSCOCO & $49.9$ & $\mathbf{50.4}$ & $48.6$ & $\mathbf{50.2}$ \\
         CelebA (F1) & $51.0$ & $\mathbf{51.8}$ & $\mathbf{57.0}$ & $53.7$\\
         LSUN & $\mathbf{56.2}$ & $55.0$ & $\mathbf{56.0}$ & $55.6$ \\
         Aircraft & $\mathbf{32.5}$ & $31.7$ & $\mathbf{37.7}$ & $33.5$ \\
         DTD & $\mathbf{30.4}$ & $28.8$ & $\mathbf{29.8}$ & $\mathbf{29.8}$ \\
         \bottomrule
    \end{tabular}
    \begin{tabular}{lrrrr}
        \toprule
         & \multicolumn{2}{c}{SimCLR} & \multicolumn{2}{c}{InstDisc}  \\
        \cmidrule(lr){2-3}
        \cmidrule(lr){4-5}
         Dataset & Expt & Ours & Expt & Ours \\
         \midrule
         MNIST & $97.1$ & $\mathbf{98.7}$ & $98.7$ & $\mathbf{98.9}$ \\
         FaMNIST & $88.3$ & $\mathbf{91.5}$ & $89.2$ & $\mathbf{91.4}$ \\
         CUBirds & $\mathbf{11.2}$ & $8.7$ & $\mathbf{13.7}$ & $9.4$ \\
         VGGFlower & $53.3$ & $\mathbf{53.6}$ & $\mathbf{61.5}$ & $54.8$ \\
         TrafficSign & $\mathbf{96.6}$ & $94.9$ & $\mathbf{98.9}$ & $94.3$ \\
         Fungi & $\mathbf{2.2}$ & $2.0$ & $\mathbf{2.6}$ & $2.1$ \\
         \bottomrule
    \end{tabular}
    \caption{\textbf{Our learned views (Ours) enable comparable transfer performance to expert views (Expt) on CIFAR-10.} Suite of transfer tasks using pretrained representations from CIFAR-10 for both the SimCLR and InstDisc pretraining setups. Numbers are percent accuracy with the exception of CelebA which is F1. FaMNIST stands for FashionMNIST.}
    \label{tab:results:cifar10}
\end{table}

We evaluate our models on CIFAR-10, as well as eleven transfer tasks including MetaDataset \citep{triantafillou2019metadataset}, MSCOCO \citep{lin2014microsoft}, MNIST \citep{lecun1998gradient}, and FashionMNIST \citep{xiao2017fashion}. We use the standard linear evaluation protocol, which trains a logistic regression on top of representations from a frozen model. We apply the same views as in pretraining, freezing the final \ours{} when using learned views; we apply no views during validation. Table \ref{tab:results:cifar10} shows our results, indicating comparable overall performance with SimCLR and InstDisc, all without the use of human-crafted view functions. This performance is noteworthy as our $\ell_1$ views cannot implement cropping-and-rescaling, which was shown to be the most important view function in \citet{chen2020simple}. We speculate that the ability of the \ours{} to implement partial masking of an image may enable a similar kind of spatial information ablation as cropping.

\subsubsection{Comparison to random $\ell_1$ noise}

Is random noise sufficient to produce domain-agnostic views? To assess how important adversarial training is to the quality of the learned representations, we perform an ablation where we generate views by adding Gaussian noise normalized to the same $\epsilon=0.05$ budget as used in the previous section. Transfer accuracy on CIFAR-10 is significantly hurt by this ablation, reaching \textbf{52.01\%} for a SimCLR model trained with random noise views compared to \textbf{84.50\%} for our method, demonstrating the importance of adversarial training to our method.

\subsubsection{The importance of inter-patch mutual information and cropping views}

Cropping-and-resizing has been identified as a crucial view function when pretraining on ImageNet \citep{chen2020simple}. However, what properties of a pretraining dataset make cropping useful? We hypothesize that such a dataset must have images whose patches have high mutual information. In other words, there must be some way for the model to identify that different patches of the same image come from the same image. While this may be true for many object or scene recognition datasets, it may be false for other important pretraining datasets, including medical or satellite imagery, where features of interest are isolated to particular parts of the image. 

\begin{figure}
\centering
\begin{subfigure}{.42\textwidth}
    \centering
    \includegraphics[width=130pt]{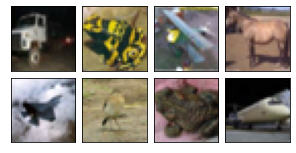}
    \caption{CIFAR-10}
    \label{subfig:cifar10}
\end{subfigure}
\begin{subfigure}{.42\textwidth}
    \centering
    \includegraphics[width=130pt]{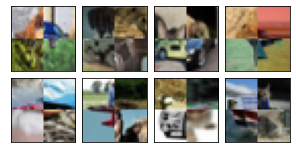}
    \caption{CIFAR-10-Corners}
    \label{subfig:cifar10-corners}
\end{subfigure}
    \caption{Our learned views are still able to yield useful information even when the inter-patch mutual information in a dataset is low, as in Figure \ref{subfig:cifar10-corners}.}
    \label{fig:collage}
\end{figure}

To investigate this hypothesis, we modify the CIFAR-10 dataset to reduce the inter-patch mutual information by replacing each 16x16 corner of the image with the corner from another image in the training dataset (see Figure \ref{fig:collage} for an example). Thus, random crops on this dataset, which we call CIFAR-10-Corners, will often contain completely unrelated information. When pretrained on CIFAR-10-Corners, expert views achieve \textbf{63.3\%} linear evaluation accuracy on the original CIFAR-10 dataset, while \ours{} views achieve \textbf{68.8\%}. This gap suggests that \ours{} views are less reliant on inter-patch mutual information than the expert views.

\subsection{Combining viewmaker and handcrafted views}
Can viewmakers improve performance in cases where some useful handcrafted views have already been identified? \citet{chen2020simple} show that views produced through cropping are significantly improved by a suite of color-based augmentations, which they argue prevents the network from relying solely on color statistics to perform the contrastive task. Here, we show that viewmaker networks also enable strong gains when added on top of cropping and horizontal flipping views when pretraining on CIFAR-10---without any domain-specific knowledge. Alone, this subset of handcrafted augmentations achieves \textbf{73.2\%} linear evaluation accuracy on CIFAR-10. Combining these views with learned viewmaker perturbations ($\eps=0.05$) achieves \textbf{83.1\%}.\footnote{We did not see additional gains from using viewmakers on top of the full, well-optimized set of SimCLR augmentations.} This suggests that viewmakers can significantly improve representation learning even in cases where some domain-specific views have already been developed.

\subsection{Robustness to common corruptions}
\label{subsec:cifar-c}

\begin{figure}
\centering
\begin{subfigure}{.45\textwidth}

    \begin{tabular}{lrrr}
    \toprule
    Views & Clean & Corrupted & Diff  \\
    \midrule
    Ours & 84.5  & 71.4 & -13.1\\
    SimCLR$^*$ & \textbf{86.2}  & 77.1 & -9.1 \\
    Combined$^*$ & \textbf{86.3} & \textbf{79.8} & \textbf{-6.5}\\
    \bottomrule
    \end{tabular}
\caption{Accuracy on CIFAR-10 and CIFAR-10-C. \\$^*$Overlap with CIFAR-10-C corruptions.}
\label{table:cifarc-acc}
\end{subfigure}
\begin{subfigure}{.45\textwidth}
    \centering
    \includegraphics[width=135pt]{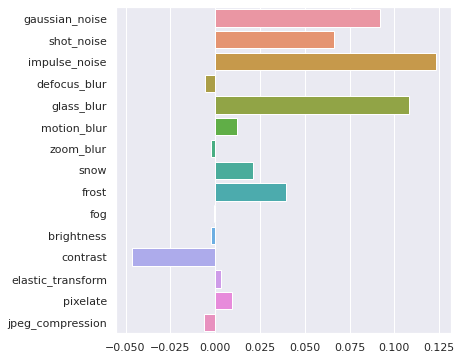}
    \caption{Accuracy gain on CIFAR-10-C by from adding our learned views atop expert views.}
    \label{subfig:vm-default-cifarc}
\end{subfigure}
    \caption{\textbf{Performance of different views on CIFAR-10-C corruptions.} Our learned views enable solid performance in the face of unseen corruptions despite not explicitly including any blurring, contrast, or brightness transformations during training, unlike the expert views. Adding our learned views on top of SimCLR yields additional gains in robust accuracy, especially on different kinds of noise corruptions and glass blurring.}
    \label{fig:cifarc}
\end{figure}

Image classification systems should behave robustly even when the data distribution is slightly different from that seen during training. Does using a \ours{} improve robustness against common types of corruptions not experienced at train time? To answer this, we evaluate both learned views, expert views, and their composition on the CIFAR-10-C dataset \citep{hendrycks2019benchmarking}, which assesses robustness to corruptions like snow, pixelation, and blurring. In this setting, corruptions are applied only at test time, evaluating whether the classification system is robust to some types of corruptions to which humans are robust.

When considering methods in isolation, SimCLR augmentations result in less of an accuracy drop from clean to corrupted data compared to our learned views, as shown in Table \ref{table:cifarc-acc}. This gap is expected, as the expert views overlap significantly with the CIFAR-10-C corruptions: both include blurring, brightness, and contrast transformations. Interestingly, however, when we train a \ours{} network while also applying expert augmentations (``Combined,'' Table \ref{table:cifarc-acc}), we can further improve the robust accuracy, with notable gains on noise and glass blur corruptions (Figure \ref{subfig:vm-default-cifarc}). This is noteworthy, as our learned views have no explicit overlap with the CIFAR-10-C corruptions, unlike the expert augmentations.\footnote{We do notice a smaller decline in contrast corruption accuracy, possibly due to interactions between changing pixel magnitudes and the $\ell_p$ constraint.} In the Combined setting, we use a distortion budget of $\epsilon=0.01$, which we find works better than $\epsilon = 0.05$, likely because combining the two augmentations at their full strength would make the learning task too difficult.

These results suggest that learned views are a promising avenue for improving robustness in self-supervised learning models.

%% file: audio.tex
\section{Speech}
Representation learning on speech data is an emerging and important research area, given the large amount of available unlabeled data and the increasing prevalence of speech-based human-computer interaction \citep{latif2020deep}. However, compared to images, there is considerably less work on self-supervised learning and data augmentations for speech data. Thus, it is a compelling setting to investigate whether \ours{} augmentations are broadly applicable across modalities.

\subsection{Self-supervised learning setup}
We adapt the contrastive learning setup from SimCLR \citep{chen2020simple}. Training proceeds largely the same as for images, but the inputs are 2D log mel spectrograms. We consider both view functions applied in the time-domain before the STFT, including noise, reverb, pitch shifts, and changes in loudness \citep{kharitonov2020data}, as well as spectral views, which involve masking or noising different parts of the spectrogram \citep{park2019specaugment}. To generate learned views, we pass the spectrogram as input to the \ours{}. We normalize the spectrogram to mean zero and variance one before passing it through the \ours{}, and do not clamp the resulting perturbed spectrogram. See the Appendix for more details. We train on the Librispeech dataset \citep{librispeech} for 200 epochs, and display some examples of learned views in the Appendix.

\subsection{Speech classification results}

\begin{table}
    \centering
    \begin{tabular}{lrrrr}
        \toprule
         & \multicolumn{2}{c}{Expert} & \multicolumn{2}{c}{Ours ($\epsilon$)}  \\
         \cmidrule(lr){2-3}
         \cmidrule(lr){4-5}
         \emph{ResNet-18, 100hr} & Time & Spec. & $0.05$ & $0.1$ \\
         \midrule
         LibriSpeech Sp. ID & $\mathbf{97.1}$ & $91.6$ & $88.3$ & $84.0$ \\
         VoxCeleb1 Sp. ID & $5.7$ & $7.8$ & $\mathbf{12.1}$ & $9.1$ \\ 
         AudioMNIST & $31.7$ & $63.9$ & $\mathbf{93.3}$ & $87.9$ \\
         Google Commands & $27.1$ & $31.9$ & $\mathbf{47.4}$ & $41.6$ \\
         Fluent Actions & $29.4$ & $32.0$ & $\mathbf{41.6}$ & $37.9$ \\
         Fluent Objects & $37.1$ & $40.3$ & $\mathbf{47.6}$ & $\mathbf{47.6}$ \\
         Fluent Locations & $59.7$ & $63.3$ & $66.5$ & $\mathbf{68.3}$ \\
         \bottomrule
    \end{tabular}
    \;
    \begin{tabular}{lrr}
        \toprule
         \emph{ResNet-50, 960hr} & Spec. & $0.05$ \\
         \midrule
         LibriSpeech Sp. ID & $\mathbf{95.9}$ & $90.0$ \\
         VoxCeleb1 Sp. ID & $8.6$ & $\mathbf{10.7}$ \\
         AudioMNIST & $80.2$ & $\mathbf{88.0}$ \\
         Google Commands & $28.3$ & $\mathbf{32.6}$ \\
         Fluent Actions & $30.5$ & $\mathbf{42.5}$ \\
         Fluent Objects & $36.2$ & $\mathbf{50.8}$ \\
         Fluent Locations & $62.0$ & $\mathbf{68.9}$ \\
         \bottomrule
    \end{tabular}
    \caption{\textbf{Our learned views significantly outperform existing views for speech transfer tasks.} Linear evaluation accuracy for SimCLR models trained on LibriSpeech. Left: ResNet-18 + Librispeech 100 hour, Right: ResNet-50 + Librispeech 960hr. ``Time'' refers to view functions applied in the time domain \citep{kharitonov2020data}, while ``Spec.'' refers to view functions applied directly to the spectrogram \citep{park2019specaugment}. $0.05$ and $0.1$ denote \ours{} distortion bounds $\epsilon$.}
    \label{tab:results:librispeech}
\end{table}

We evaluate on three speech classification datasets: Fluent Speech Commands \citep{lugosch2019speech}, Google Speech Commands \citep{warden2018speech}, and spoken digit classification \citep{becker2018interpreting}, as well as speaker classification on VoxCeleb \citep{nagrani2017voxceleb} and Librispeech \citep{librispeech}, all using the linear evaluation protocol for 100 epochs. In Table \ref{tab:results:librispeech}, we report results with both the same distortion budget $\epsilon = 0.05$ as in the image domain, as well as a larger $\epsilon = 0.1$, for comparison. Both versions significantly outperform the preexisting waveform and spectral augmentations, with a $+9$ percentage point improvement on average for the ResNet-18 ($\epsilon=0.05$) viewmaker model over the best expert views. The gains for real-world tasks such as command identification are compelling. One notable exception is the task of LibriSpeech speaker identification. Since LibriSpeech is the same dataset the model was pretrained on, and this effect is not replicated on VoxCeleb1, the other speaker classification dataset, we suspect the model may be picking up on dataset-specific artifacts (e.g. background noise, microphone type) which may make the speaker ID task artificially easier. An interesting possibility is that the worse performance of \ours{} views may result from the model being able to identify and ablate such spurious correlations in the spectrograms.

%% file: imu.tex
\section{Wearable sensor data}

To further validate that our method for learning views is useful across different modalities, we consider time-series data from wearable sensors. Wearable sensor data has a broad range of applications, including health care, entertainment, and education \citep{lara2012survey}.  We specifically consider whether \ours{} views improve representation learning for the task of human activity recognition (HAR), for example identifying whether a user is jumping rope, running, or cycling. 

\subsection{Self-supervised learning setup}

We consider the Pamap2 dataset \citep{reiss2012introducing}, a dataset of 12 different activities performed by 9 participants. Each activity contains 52 different time series, including heart rate, accelerometer, gyroscope, and magnetometer data collected from sensors on the ankle, hand, and chest (all sampled at 100Hz, except heart rate, which is sampled at approximately 9Hz). We linearly interpolate missing data, then take random 10s windows from subject recordings, using the same train/validation/test splits as prior work \citep{moya2018convolutional}. To create inputs for our model, we generate a multi-channel image composed of one 32x32 log spectrogram for each sensor time-series window. Unlike speech data, we do not use the mel scale when generating the spectrogram. We then normalize the training and validation datasets by subtracting the mean and then dividing by the standard deviation of the training dataset.  

We train with both our learned views and the spectral views \citep{park2019specaugment} that were most successful in the speech domain (for multi-channel spectral masking, we apply the same randomly chosen mask to all channels). We also compare against a variant of these views with spectrogram noise removed, which we find improves this baseline's performance.

\subsection{Sensor-based activity recognition results}

We train a linear classifier on the frozen encoder representations for 50 epochs, reporting accuracy on the validation set. We sample 10k examples for each training epoch and 50k examples for validation. Our views significantly outperform spectral masking by 12.8 percentage points when using the same $\epsilon=0.05$ as image and speech, and by 16.7 points when using a larger $\epsilon=0.5$ (Table \ref{tab:results:imu}). We also find that a broad range of distortion budgets produces useful representations, although overly-aggressive budgets prevent learning (Table \ref{tab:results:imu}). These results provide further evidence that our method for learning views has broad applicability across different domains.

\begin{table}
    \centering
    \begin{tabular}{lrrrrrrr}
        \toprule
         & \multicolumn{2}{c}{Spectral} & \multicolumn{5}{c}{Ours ($\epsilon$)}  \\
         \cmidrule(lr){2-3}
         \cmidrule(lr){4-8}        
         Dataset & With Noise & Without Noise & $0.02$ & $0.05$ & $0.2$ & $0.5$  & $2.0$ \\
         \midrule
         Pamap2 & $71.0$ & $74.6$ & $83.0$ & $87.4$ & $88.6$ & $\mathbf{91.3}$ & $9.1$ \\
         \bottomrule
    \end{tabular}
    \caption{\textbf{Our learned views significantly outperform existing views for activity recognition on wearable sensor data.} Our method learns superior representations across a large range of distortion budgets $\epsilon$, although budgets that are too strong prevent learning.  Linear evaluation accuracy for  ResNet18 models trained on Pamap2 with SimCLR. ``Spectral'' refers to view functions applied directly to the spectrogram \citep{park2019specaugment}.}
    \label{tab:results:imu}
\end{table}

\subsection{Semi-supervised experiments}

An especially important setting for self-supervised learning is domains where labeled data is scarce or costly to acquire. Here, we show that our method can enable strong performance when labels for only a single participant (Participant 1) out of seven are available. We compare simple supervised learning on Participant 1's labels against linear evaluation of our best pretrained model, which was trained on unlabeled data from all 7 participants. The model architectures and training procedures are otherwise identical to the previous section. As Figure \ref{tab:results:imu-semisup} shows, pretraining with our method on unlabeled data enables significant gains over pure supervised learning when data is scarce, and even slightly outperforms the hand-crafted views trained on all 7 participants (cf.~Table \ref{tab:results:imu}).

\begin{table}[h]
    \centering
    \begin{tabular}{lrrrr}
        \toprule
         & \multicolumn{2}{c}{Supervised Learning} & \multicolumn{2}{c}{Pretrain (Ours) \& Transfer}  \\
        \cmidrule(lr){2-3}
        \cmidrule(lr){4-5}
         Dataset & 1 Participant & 7 Participants & 1 Participant & 7 Participants \\
         \midrule
         Pamap2 & $58.3$ & $97.1$ & $75.1$ & $91.3$ \\
         \bottomrule
    \end{tabular}
    \caption{\textbf{Our method enables superior results in a semi-supervised setting where labels for data from only one participant are available.} Validation accuracy for activity recognition on Pamap2. Supervised Learning refers to training a randomly initialized model on the labeled data until convergence. Pretrain \& Transfer refers to training a linear classifier off of the best pretrained model above. 1 or 7 Participants refers to the number of participants comprising the training set.}
    \label{tab:results:imu-semisup}
\end{table}

%% file: conclusion.tex
\section{Conclusion}

We introduce a method for learning views for unsupervised learning, demonstrating its effectiveness through strong performance on image, speech, and wearable sensor modalities. Our novel generative model---\ours{} networks---enables us to efficiently learn views as \emph{part of} the representation learning process, as opposed to relying on domain-specific knowledge or costly trial and error. There are many interesting avenues for future work. For example, while the $\ell_1$ constraint is simple by design, there may be other kinds of constraints that enable richer spaces of views and better performance. In addition, \ours{} networks may find use in supervised learning, for the purposes of data augmentation or improving robustness. Finally, it is interesting to consider what happens as the \ours{} networks increase in size: do we see performance gains or robustness-accuracy trade-offs \citep{raghunathan2019adversarial}? Ultimately, our work is a step towards more general self-supervised algorithms capable of pretraining on arbitrary data and domains.

%% file: ack.tex
\subsubsection*{Acknowledgements}
We would like to thank Dan Yamins, Chengxu Zhuang, Shyamal Buch, Jesse Mu, Jared Davis, Aditi Raghunathan, Pranav Rajpurkar, Margalit Glasgow, and Jesse Michel for useful discussions and comments on drafts. AT is supported by an Open Phil AI Fellowship. MW is supported by the Stanford Interdisciplinary Graduate Fellowship as the Karr Family Fellow.

%% file: appendix.tex
\section{Additional experimental details}

\subsection{Image experiments}
The primary image experiments compare SimCLR and the instance discrimination method from  \citep{wu2018unsupervised} (henceforth InstDisc) with and without the \ours{} on a suite of transfer datasets. 

For pretraining, we use a modified ResNet-18 encoder specialized for smaller inputs, such as CIFAR-10.\footnote{Adapted from \url{https://github.com/kuangliu/pytorch-cifar/blob/master/models/resnet.py}} We found this to be crucial for performance across all models when working with smaller input images of 32x32 pixels. We use an embedding dimension of size 128 and do not use an additional projection head as in \citet{chen2020big}. For the SimCLR objective, we use a temperature of 0.07. For the InstDisc objective, we use 4096 negative samples from the memory bank and an update rate of 0.5. We optimize using SGD with batch size 256, learning rate 0.03, momentum 0.9, and weight decay 1e-4 for 200 epochs with no learning rate dropping, which we found to hurt performance in CIFAR-10. 

For the \ours{}, we adapt the style transfer network from \citet{10.1007/978-3-319-46475-6_43}, using a PyTorch implementation,\footnote{\url{https://github.com/pytorch/examples/tree/master/fast_neural_style}} but use three residual blocks instead of five, which we found did not hurt performance despite the reduced computation.
To add stochasticity, we concatenate a uniform random noise channel to the input and the activations before each residual block. 

Additionally, we performed preliminary experiments with a U-Net architecture \citep{ronneberger2015u} for the \ours{} but found significantly worse performance. We leave a more in-depth investigation of the role of architecture and model size in the effectiveness of the viewmaker.

During transfer (linear evaluation), we use the pre-pooling features after the last convolutional layer of the modified ResNet-18, totaling 512*4*4 dimensions. We load the parameters from the final iteration of pretraining. We optimize a logistic regression model with the frozen ResNet-18 model using SGD with learning rate 0.01, momentum 0.9, weight decay 0, batch size 128 for 100 epochs. We drop the learning rate by a factor of 10 on epochs 60 and 80. We preprocess the training and validation datasets by subtracting and dividing by the mean and standard deviation of the training dataset, respectively. For models trained with a \ours{} network, we load and freeze the final \ours{} checkpoint to supply augmentations during transfer training. Otherwise, we use the same expert views used during pretraining.

The CIFAR-10-Corners experiments were conducted in the same way, except that the transfer task is the original CIFAR-10 dataset. 

For the robustness experiments on CIFAR-10-C, the final transfer model trained on CIFAR-10 was evaluated without further training on the CIFAR-10-C dataset.

\subsection{Speech experiments}

The setup for the speech experiments is almost identical to images. The primary distinction is in preprocessing the data. In our experiments, pretraining is done on two splits of LibriSpeech: a smaller set containing 100 hours of audio and a larger set containing 960 hours of audio. Each instance is a raw waveform. We pick a maximum limit of 150k frames and truncate waveforms containing more frames. We randomly pick whether to truncate the beginning or end of the waveform during training, whereas for evaluation, we always truncate the end. Next, we compute log mel spectrograms on the truncated waveforms as the input to our encoder. For 100hr LibriSpeech, we use a hop length of 2360 and set the FFT window length to be 64, resulting in a 64x64 tensor. For the 960hr LibriSpeech, we wanted to show our method generalizes to larger inputs, so we use a hop length of 672 with an FFT window length of 112 for a tensor of size 112x112. Finally, we log the spectrogram by squaring it and converting power to decibels.

For expert views, we consider both a method that applies views directly to the waveforms \citep{kharitonov2020data} and a method that does so on the resulting spectrograms \citep{park2019specaugment}. For the former, we use code from the \textsc{nlpaug} library\footnote{\url{https://github.com/makcedward/nlpaug}} to take a random contiguous crop of the waveform with scale (0.08,1.0) and add Gaussian noise with scale 1. We randomly mask contiguous segments on the horizontal (frequency) and vertical (time) axes for the latter. 

To do this, we also use the \textsc{nlpaug} library and employ the \textsc{FrequencyMaskingAug} and \textsc{TimeMaskingAug} functions with \textsc{mask\_factor} set to 40. Having done this, we are left with a 1x64x64 tensor for the 100-hour dataset or a 1x112x112 tensor for the 960-hour dataset. For the former, we use a standard ResNet-18; pretraining and transfer use the same hyperparameters. For the latter, we use a standard ResNet-50 encoder with an MLP projection head with a hidden dimension of 2048. We use \textsc{torchvision} implementations \citep{paszke2019pytorch}. We still use the pre-pooling features for transfer in this setting as we found better performance than using post-pooling features. Otherwise, hyperparameters are identical to the 100-hour setting (and the CIFAR-10 setting). 

\def\UrlBigBreaks{\do\/\do-\do:}
For each transfer dataset, we convert waveforms to normalized spectrograms in the same manner as just described. The AudioMNIST dataset was downloaded from \url{https://github.com/soerenab/AudioMNIST}; Google Speech Commands was downloaded from \url{https://ai.googleblog.com/2017/08/launching-speech-commands-dataset.html}; Fluent Speech Commands was downloaded from \url{https://fluent.ai/fluent-speech-commands-a-dataset-for-spoken-language-understanding-research}; VoxCeleb1 was downloaded from \url{http://www.robots.ox.ac.uk/~vgg/data/voxceleb/vox1.html} (we use version 1 of the corpus). Each transfer dataset was again normalized using the training split's mean and standard deviation.

\subsection{Wearable sensor experiments}

The experimental paradigm for wearable sensor data largely follows that for speech. To generate an example, we randomly sample a subject (from the correct training split) and activity; we next randomly sample a contiguous 10s frame, linearly interpolating missing data. We generate spectrograms for each of the 52 sensors without Mel scaling, using 63 FFT bins, a hop length of 32, and a power of 2, then take the logarithm after adding 1e-6 for numerical stability. This process yields 52 32x32 spectrograms, which we treat as different channels, giving a tensor of shape [52, 32, 32]. We then normalize the spectrograms by subtracting and dividing by the mean and standard deviation of 10k samples from the training set. We again use the modified ResNet-18 encoder for smaller inputs here.

\subsection{Training costs}

We train all models on single NVIDIA Titan XP GPUs. On our system, training with a \ours{} network roughly increased training time by 50\% and GPU memory utilization by 100\%.

\subsection{Frameworks}

We make use of PyTorch \citep{paszke2019pytorch}, PyTorch Lightning \citep{falcon2019pytorch}, and Weights \& Biases \citep{wandb} for our experiments.

\section{Additional generated views}

\subsection{CIFAR-10 views}
\begin{figure}
    \centering
    \includegraphics[width=0.37\linewidth]{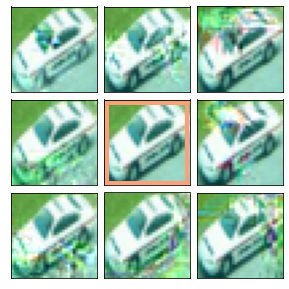}
    \includegraphics[width=0.37\linewidth]{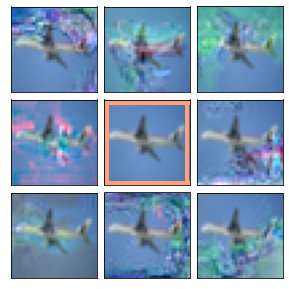}
    \includegraphics[width=0.37\linewidth]{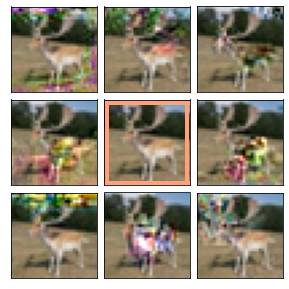}
    \includegraphics[width=0.37\linewidth]{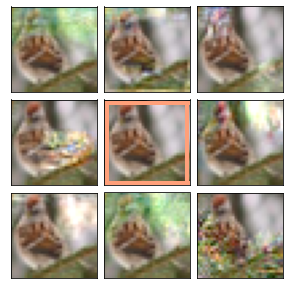}
    \includegraphics[width=0.37\linewidth]{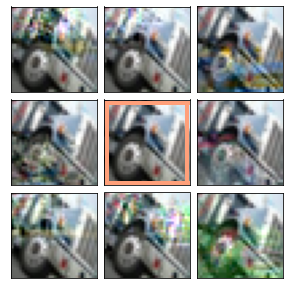}
    \includegraphics[width=0.37\linewidth]{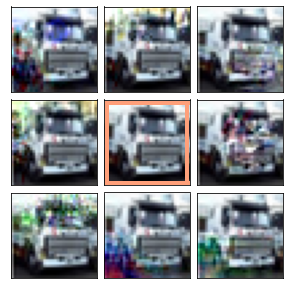}
    \caption{Learned views for random CIFAR-10 examples. Original image shown in center, with pink border. Distortion budget is $\epsilon=0.05$. }
    \label{fig:cifar-views-more}
\end{figure}

\begin{figure}
    \centering
    \includegraphics[width=0.37\linewidth]{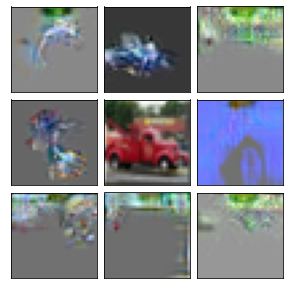}
    \includegraphics[width=0.37\linewidth]{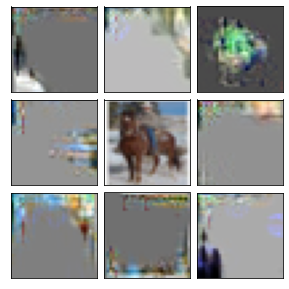}
    \includegraphics[width=0.37\linewidth]{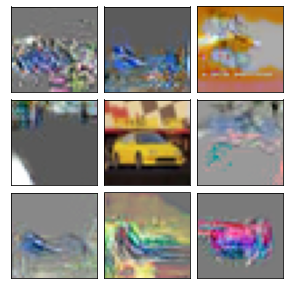}
    \includegraphics[width=0.37\linewidth]{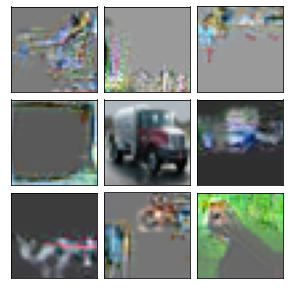}
    \includegraphics[width=0.37\linewidth]{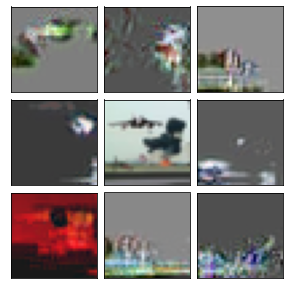}
    \includegraphics[width=0.37\linewidth]{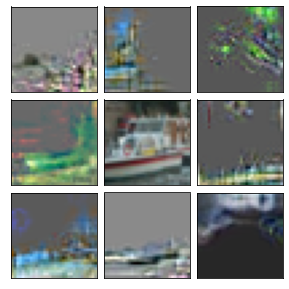}
    \caption{Difference between random CIFAR-10 examples and their \ours{} views. Original image shown in center, diffs shown on perimeter. Diffs linearly rescaled to [0, 1]. Distortion budget is $\epsilon=0.05$.}
    \label{fig:cifar-views-deltas}
\end{figure}

We visualize more views for CIFAR in Figure \ref{fig:cifar-views-more}. We also visualize the difference between examples and their views (rescaled to [0,1]) in Figure \ref{fig:cifar-views-deltas}. These figures further demonstrate the complexity, diversity, and input dependence of the \ours{} views.

\subsubsection{Applying perturbations in the frequency domain}
\begin{figure}
    \centering
    \includegraphics[width=0.37\linewidth]{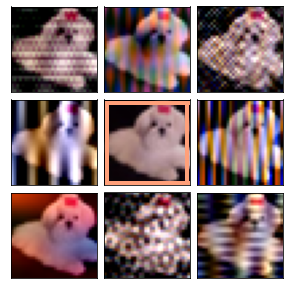}
    \includegraphics[width=0.37\linewidth]{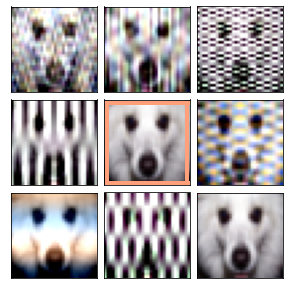}
    \includegraphics[width=0.37\linewidth]{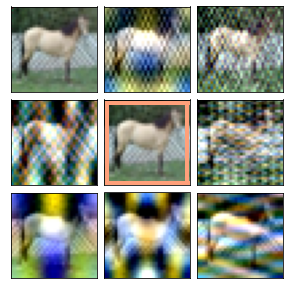}
    \includegraphics[width=0.37\linewidth]{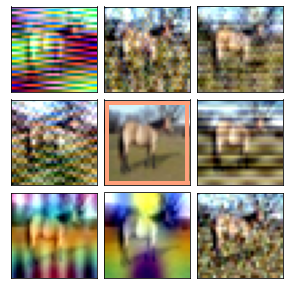}
    \includegraphics[width=0.37\linewidth]{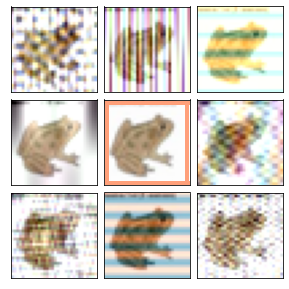}
    \includegraphics[width=0.37\linewidth]{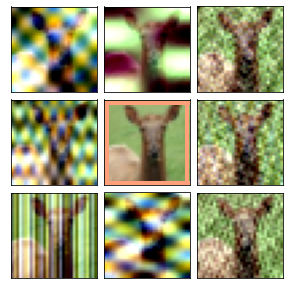}
    \caption{Learned views for random CIFAR-10 examples with perturbation applied in frequency domain. Original image shown in center, with pink border. Distortion budget is $\epsilon=1.0$.}
    \label{fig:cifar-spectral-views}
\end{figure}

Are there other natural ways of generating perturbations with bounded complexity? One other technique we considered was applying views in the frequency domain. Specifically, we apply a Discrete Cosine Transform \citep[DCT]{ahmed1974discrete} before applying the $\ell_1$-bounded perturbation, then apply the inverse DCT to obtain an image in the original domain. We use a PyTorch library\footnote{\url{https://github.com/zh217/torch-dct}} to compute the DCT, which is a differentiable transform. After a coarse hyperparameter search, we achieved the best results with $\epsilon = 1.0$: \textbf{74.4\%} linear evaluation accuracy on CIFAR-10, much lower than our other models. However, the views are still illustrative, and we show some examples in Figure \ref{fig:cifar-spectral-views}.

\subsection{Librispeech views}

\begin{figure}
    \centering
    \includegraphics[width=0.37\linewidth]{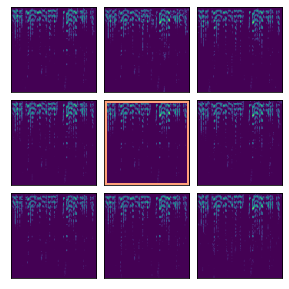}
    \includegraphics[width=0.37\linewidth]{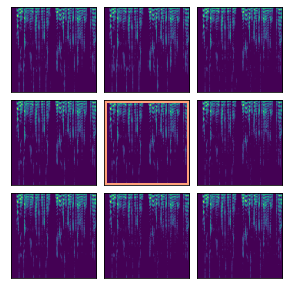}
    \includegraphics[width=0.37\linewidth]{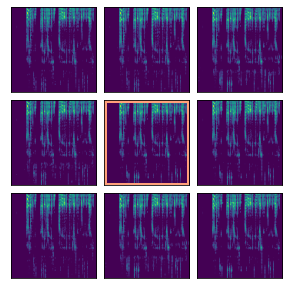}
    \includegraphics[width=0.37\linewidth]{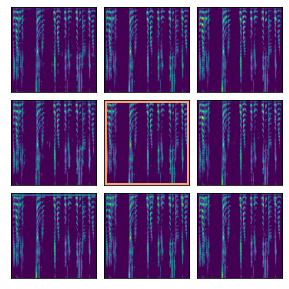}
    \includegraphics[width=0.37\linewidth]{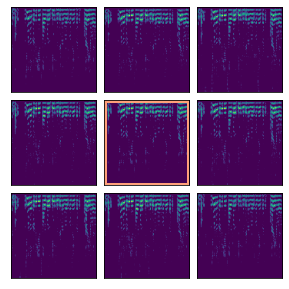}
    \includegraphics[width=0.37\linewidth]{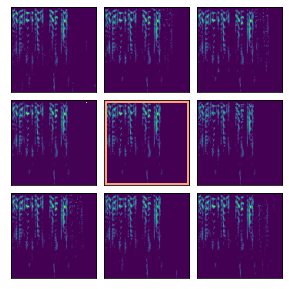}
    \includegraphics[width=0.37\linewidth]{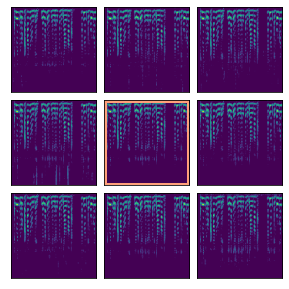}
    \includegraphics[width=0.37\linewidth]{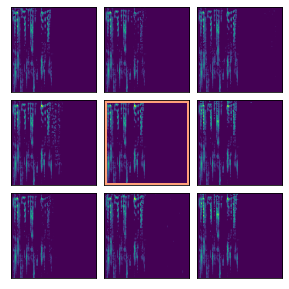}
    \caption{Examples of learned views for random Librispeech spectrograms. Original image shown in center, with pink border. Variations are subtle---best viewed at high magnification. Color scale endpoints set to minimum and maximum of original image. Spectrograms are 64x64 log mel spectrograms from LibriSpeech 100 hours. Distortion budget is $\epsilon=0.05$. }
    \label{fig:librispeech-views}
\end{figure}

\begin{figure}
    \centering
    \includegraphics[width=0.37\linewidth]{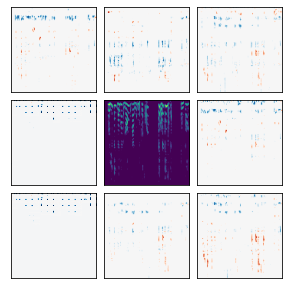}
    \includegraphics[width=0.37\linewidth]{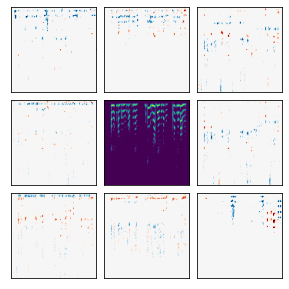}
    \includegraphics[width=0.37\linewidth]{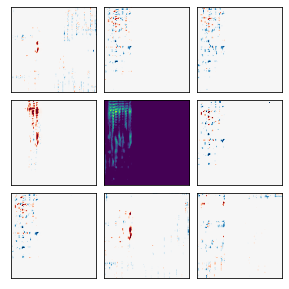}
    \includegraphics[width=0.37\linewidth]{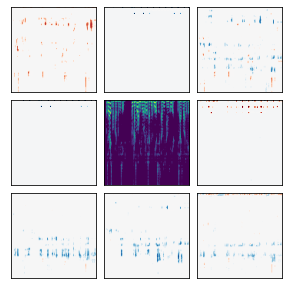}
    \includegraphics[width=0.37\linewidth]{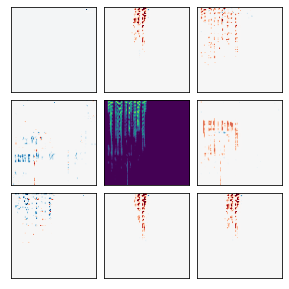}
    \includegraphics[width=0.37\linewidth]{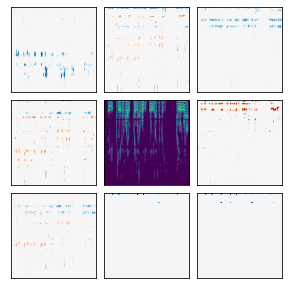}
    \caption{Difference between random LibriSpeech spectrograms and their \ours{} views. Original spectrogram shown in center, diffs shown on perimeter. Color scale endpoints set to -2.5 (red) to +2.5 (blue), although some values exceed these endpoints. Spectrograms are 64x64 log mel spectrograms from LibriSpeech 960 hours. Distortion budget is $\epsilon=0.05$.}
    \label{fig:librispeech-views-deltas}
\end{figure}

We visualize some views for random LibriSpeech spectrograms in Figure \ref{fig:librispeech-views}, as well as showing deltas between spectrograms and views in Figure \ref{fig:librispeech-views-deltas}. The figures show how the \ours{} applies a variety of kinds of perturbations across the entire spectrogram.

\subsection{Wearable sensor views}

\begin{figure}
    \centering
    \includegraphics[width=0.14\linewidth]{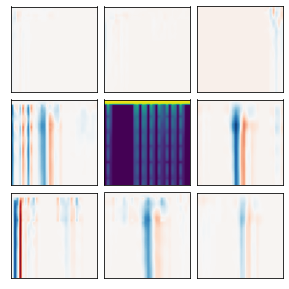}
    \includegraphics[width=0.14\linewidth]{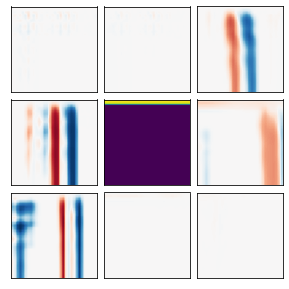}
    \includegraphics[width=0.14\linewidth]{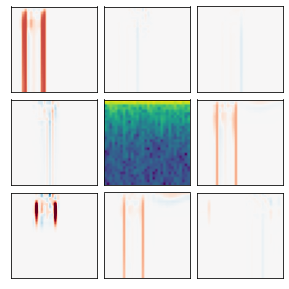}
    \includegraphics[width=0.14\linewidth]{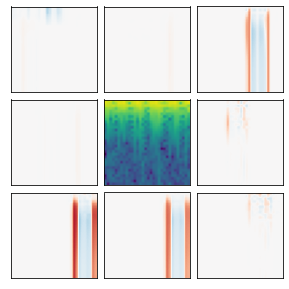}
    \includegraphics[width=0.14\linewidth]{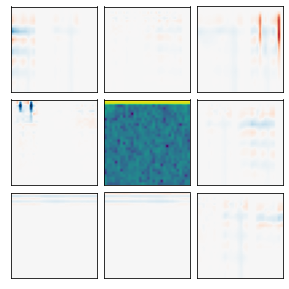}
    \includegraphics[width=0.14\linewidth]{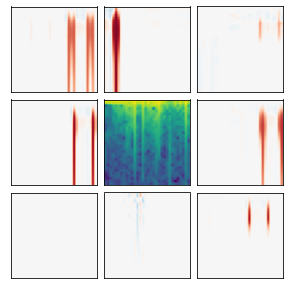}
    \includegraphics[width=0.14\linewidth]{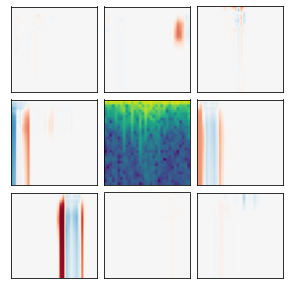}
    \includegraphics[width=0.14\linewidth]{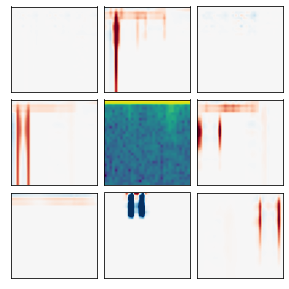}
    \includegraphics[width=0.14\linewidth]{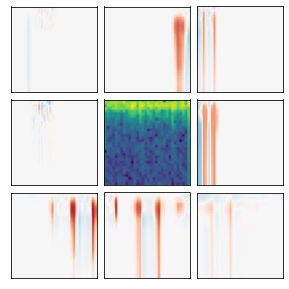}
    \includegraphics[width=0.14\linewidth]{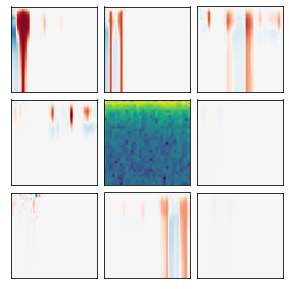}
    \includegraphics[width=0.14\linewidth]{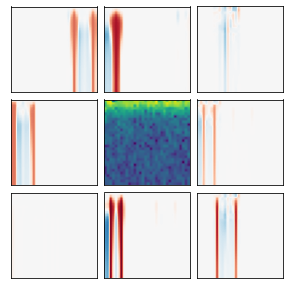}
    \includegraphics[width=0.14\linewidth]{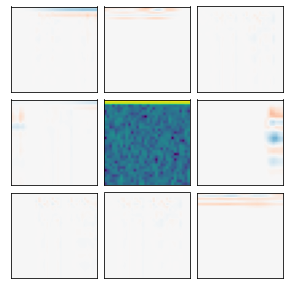}
    \includegraphics[width=0.14\linewidth]{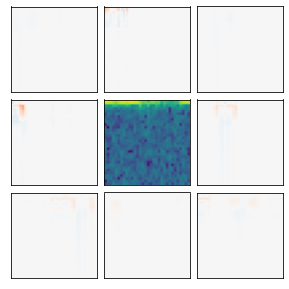}
    \includegraphics[width=0.14\linewidth]{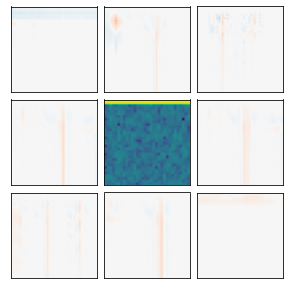}
    \includegraphics[width=0.14\linewidth]{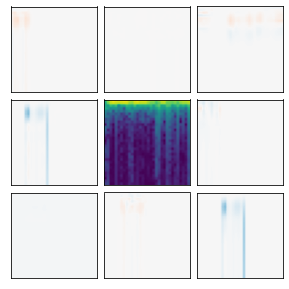}
    \includegraphics[width=0.14\linewidth]{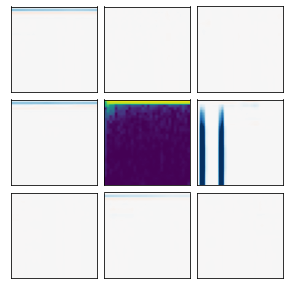}
    \includegraphics[width=0.14\linewidth]{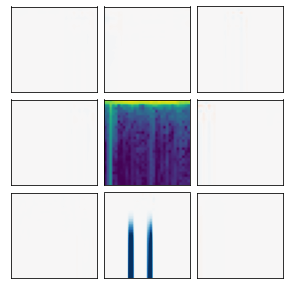}
    \includegraphics[width=0.14\linewidth]{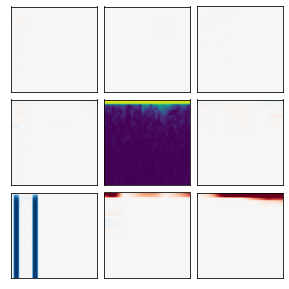}
    \includegraphics[width=0.14\linewidth]{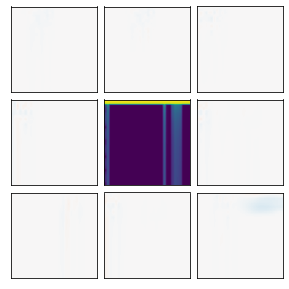}
    \includegraphics[width=0.14\linewidth]{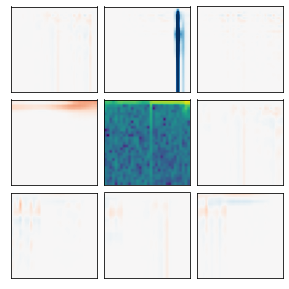}
    \includegraphics[width=0.14\linewidth]{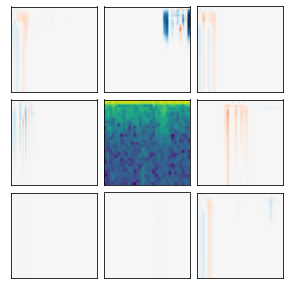}
    \includegraphics[width=0.14\linewidth]{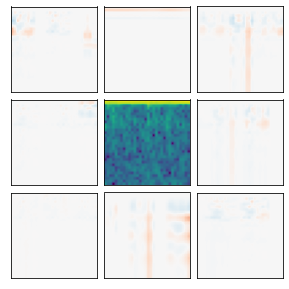}
    \includegraphics[width=0.14\linewidth]{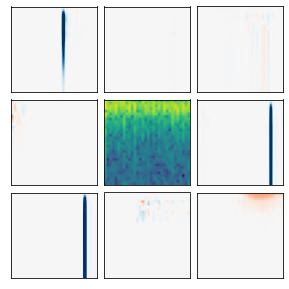}
    \includegraphics[width=0.14\linewidth]{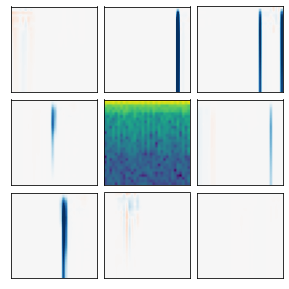}
    \includegraphics[width=0.14\linewidth]{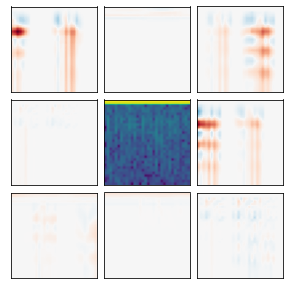}
    \includegraphics[width=0.14\linewidth]{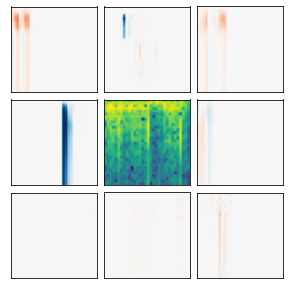}
    \includegraphics[width=0.14\linewidth]{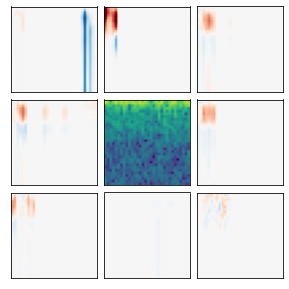}
    \includegraphics[width=0.14\linewidth]{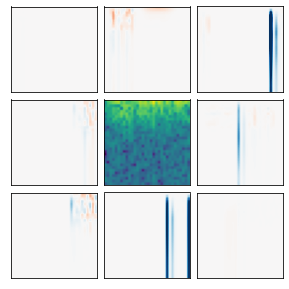}
    \includegraphics[width=0.14\linewidth]{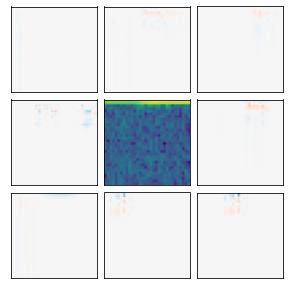}
    \includegraphics[width=0.14\linewidth]{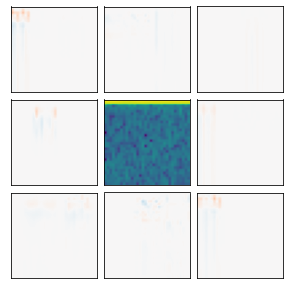}
    \includegraphics[width=0.14\linewidth]{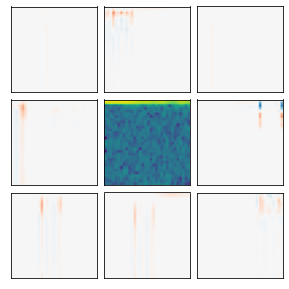}
    \includegraphics[width=0.14\linewidth]{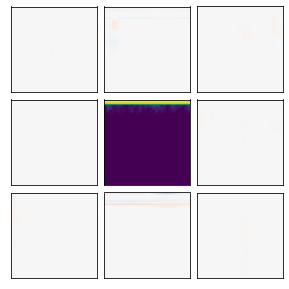}
    \includegraphics[width=0.14\linewidth]{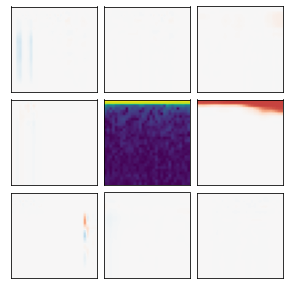}
    \includegraphics[width=0.14\linewidth]{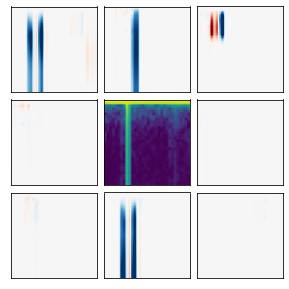}
    \includegraphics[width=0.14\linewidth]{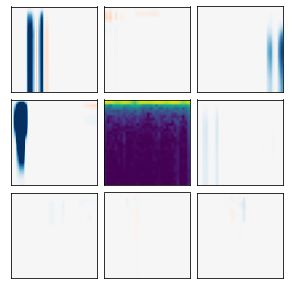}
    \includegraphics[width=0.14\linewidth]{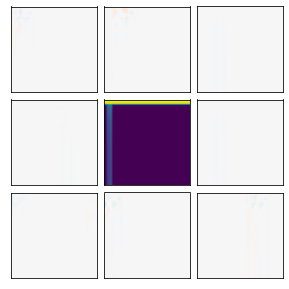}
    \includegraphics[width=0.14\linewidth]{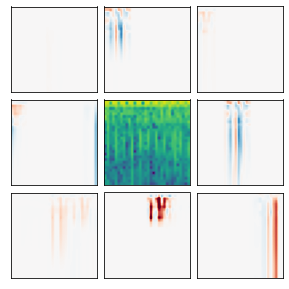}
    \includegraphics[width=0.14\linewidth]{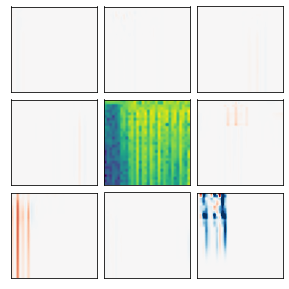}
    \includegraphics[width=0.14\linewidth]{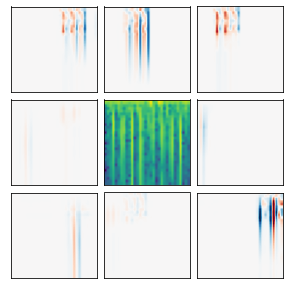}
    \includegraphics[width=0.14\linewidth]{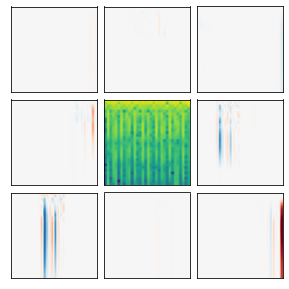}
    \includegraphics[width=0.14\linewidth]{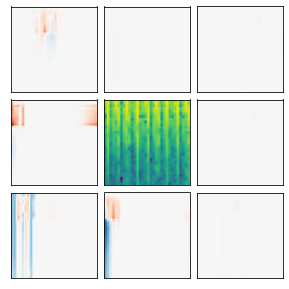}
    \includegraphics[width=0.14\linewidth]{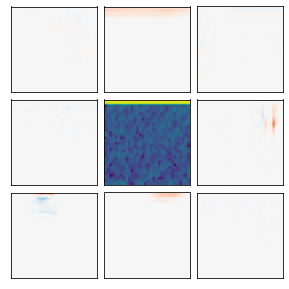}
    \includegraphics[width=0.14\linewidth]{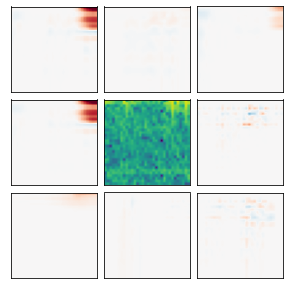}
    \includegraphics[width=0.14\linewidth]{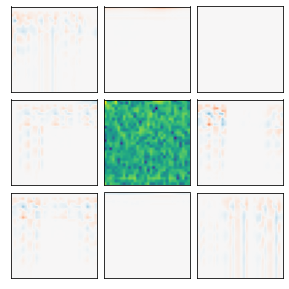}
    \includegraphics[width=0.14\linewidth]{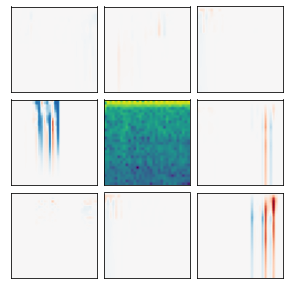}
    \includegraphics[width=0.14\linewidth]{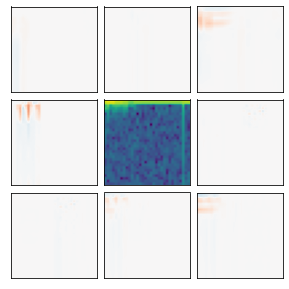}
    \includegraphics[width=0.14\linewidth]{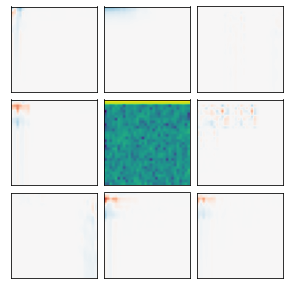}
    \includegraphics[width=0.14\linewidth]{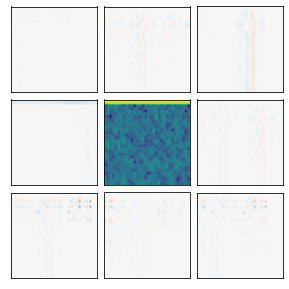}
    \includegraphics[width=0.14\linewidth]{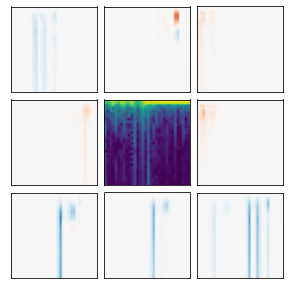}
    \includegraphics[width=0.14\linewidth]{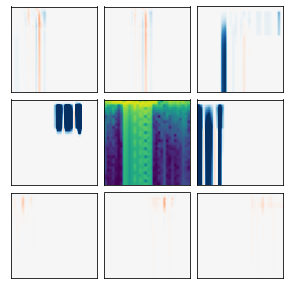}
    \includegraphics[width=0.14\linewidth]{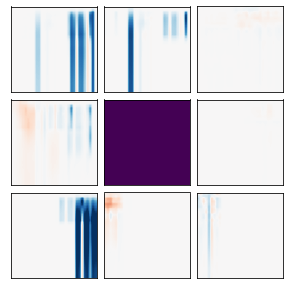}
    \includegraphics[width=0.14\linewidth]{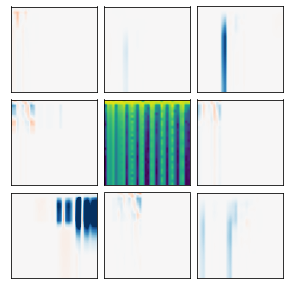}
    \caption{Difference between random Pamap2 spectrograms and their \ours{} views. Original spectrogram shown in center, diffs shown on perimeter. Each 3x3 panel shows data from a different example and sensor. Color scale endpoints set to -2 (red) to +2 (blue), although some values exceed these endpoints. Distortion budget is $\epsilon=0.05$.}
    \label{fig:sensor-views-deltas}
\end{figure}

We visualize deltas between Pamap2 spectrograms and their views in Figure \ref{fig:sensor-views-deltas}. Each 3x3 panel shows data and views from a different sensor and example.

\subsection{Stability across random seeds}

While instability has been reported as a common issue when training GANs \citep{goodfellow2016nips}, we encountered few optimization difficulties training viewmakers. To empirically demonstrate the stability of our approach across random seeds, we report the average and standard deviation of transfer accuracy across three pretraining and transfer runs for different datasets. The experimental setup is identical to the results presented in Table \ref{tab:results:cifar10}, and the random seeds vary across both pretraining and transfer. Table \ref{table:stability} shows that the observed standard deviations are small, lying within a percentage point in all but one case.

\begin{table}
\centering
\begin{tabular}{lcc}
\toprule
Dataset & Ours & Expert  \\
\midrule
Aircraft & 32.0 (0.7) & 32.0 (0.6) \\
Birds & 8.7 (0.3) & 10.9 (0.3) \\
DTD & 27.8 (0.9) & 30.4 (1.1) \\
FaMNIST & 91.0 (0.4) & 88.5 (0.2) \\
MNIST & 98.8 (0.1) & 97.1 (0.0) \\
Traffic & 94.8 (1.0) & 96.7 (0.3) \\
Flower & 50.6 (2.6) & 53.2 (0.4) \\
\bottomrule
\end{tabular}
\caption{\textbf{Stability of viewmaker networks across random seeds.} Linear evaluation accuracy and standard deviation for three random seeds, where the seed varies across both pretraining and transfer. Experimental setup is identical to that of Table \ref{tab:results:cifar10}.}
\label{table:stability}
\end{table}

\subsection{Top-5 accuracy for VoxCeleb Speaker Identification}

We also present Top-5 accuracies for VoxCeleb speaker identification in Table \ref{tab:voxceleb-top5}, along with with Top-1 accuracies for comparison.

\begin{table}
    \centering
    \begin{tabular}{lrrrr}
        \toprule
         & \multicolumn{2}{c}{Expert} & \multicolumn{2}{c}{Ours ($\epsilon$)}  \\
         \cmidrule(lr){2-3}
         \cmidrule(lr){4-5}
         \emph{ResNet-18, 100hr} & Time & Spec. & $0.05$ & $0.1$ \\
         \midrule
         Top-1 Accuracy & $\mathbf{97.1}$ & $91.6$ & $88.3$ & $84.0$ \\
         Top-5 Accuracy & $5.7$ & $7.8$ & $\mathbf{12.1}$ & $9.1$ \\ 
         \bottomrule
    \end{tabular}
    \;
    \begin{tabular}{lrr}
        \toprule
         \emph{ResNet-50, 960hr} & Spec. & $0.05$ \\
         \midrule
         Top-1 Accuracy  & $\mathbf{97.1}$ & $91.6$ \\
         Top-5 Accuracy  & $5.7$ & $7.8$ \\ 
         \bottomrule
    \end{tabular}
    \caption{VoxCeleb speaker identification linear evaluation accuracy. Experimental setup identical to Table \ref{tab:results:librispeech}.}
    \label{tab:voxceleb-top5}
\end{table}